\documentclass[letterpaper, 10 pt, conference]{IEEEtran}  

\IEEEoverridecommandlockouts

\usepackage{cite}
\usepackage{amsmath,amssymb,amsfonts}
\usepackage{graphicx}
\usepackage{textcomp}
\usepackage{xcolor}
\def\BibTeX{{\rm B\kern-.05em{\sc i\kern-.025em b}\kern-.08em
    T\kern-.1667em\lower.7ex\hbox{E}\kern-.125emX}}

\usepackage{multicol}

\usepackage[hidelinks]{hyperref}
\definecolor{darkred}{RGB}{150,0,0}
\definecolor{darkgreen}{RGB}{0,150,0}
\definecolor{darkblue}{RGB}{0,0,150}
\hypersetup{colorlinks=true, linkcolor=red, citecolor=blue, urlcolor=darkblue}
\usepackage{booktabs}

\usepackage{blindtext}
\usepackage{algorithm}
\usepackage{algpseudocode}
\usepackage{color}
\usepackage{array}
\usepackage[makeroom]{cancel}
\usepackage{changes}
\usepackage{parskip}
\usepackage{comment}
\usepackage{qcircuit}
\usepackage{braket}
\usepackage{bbm}
\usepackage{bm}
\usepackage{multirow}
\usepackage{mathrsfs}
\usepackage{url}            
\usepackage{amsfonts}       
\usepackage{nicefrac}       
\usepackage{microtype}      
\usepackage[capitalize]{cleveref}

\usepackage{enumitem}

\newtheorem{problem}{Problem}

\title{\LARGE \bf
Agentic AI for Trip Planning Optimization Application
}

\author{Tiejin Chen$^{1,2}$, Ahmadreza Moradipari$^1$, Kyungtae Han$^1$, Hua Wei$^2$, and Nejib Ammar$^1$
\thanks{$^1$ Toyota Motor North America R\&D, InfoTech Labs, USA. The work was done while Tiejin Chen was an intern at Toyota InfoTech Lab. $^2$ Arizona State University.
        {\tt\small \{ahmadreza.moradipari, kyungtae.han, nejib.ammar\}@toyota.com}
        }%
}

\begin{document}

\maketitle

\begin{abstract}
Trip planning for intelligent vehicles increasingly requires selecting optimal routes rather than merely producing feasible itineraries, as interacting factors such as travel time, energy consumption, and traffic conditions directly affect plan quality. Yet existing systems are largely designed for feasibility-oriented planning, and current benchmarks provide only reference answers without ground truth, preventing objective evaluation of optimization performance. In our paper, we address these limitations with an agentic AI framework that enables dynamic refinement through an orchestration agent coordinating specialized agents for traffic, charging, and points of interest, and with the Trip-planning Optimization Problems Dataset, which supplies definitive optimal solutions and category-level task structure for fine-grained analysis. Experiments show that our system achieves 77.4\% accuracy on the TOP Benchmark, significantly outperforming single-agent and workflow-based multi-agent baselines, demonstrating the importance of orchestrated agentic reasoning for robust trip planning optimization.
\end{abstract}

\section{Introduction}


Trip planning has long been approached as the task of producing a feasible itinerary that satisfies a user’s high-level preferences. In this conventional formulation, a system is expected to recommend plausible destinations, arrange visit sequences, or outline an activity plan that aligns with user intent~\cite{kim2015use,vansteenwegen2010trip}. When trip planning is considered in the context of intelligent and connected vehicles, a feasibility-only perspective becomes insufficient. A vehicle must account for dynamic and interacting factors such as travel time, traffic patterns, energy consumption, and charging availability, all of which jointly influence the actual quality of a plan. Even a route that is feasible in principle may lead to avoidable delays or inefficient energy use are taken into account. These characteristics naturally shift trip planning from producing any feasible plan to selecting an optimized one that performs well under practical constraints. 


However, there exist two fundamental gaps that prevent current research from supporting trip planning optimization. The first gap is a methodology gap that concerns the design of planning systems. Earlier approaches to automated trip planning relied on classical pathfinder algorithms~\cite{zha2022revisiting}, which are effective for computing shortest paths but are not designed to capture user preferences. Manual route construction remains widespread~\cite{xiang2015adapting}, yet it is time-consuming and often leads to suboptimal choices when users must process large amounts of information. More recent systems based on large language models introduced a new paradigm for interactive travel assistance and can interpret natural language queries together with user-specified constraints, but they still aim to produce feasible itineraries rather than to identify the best possible plan under competing objectives~\cite{chen2024travelagent}. As a result, existing methods are not equipped to perform trip planning optimization, leaving a clear methodological gap.

The second gap is an evaluation gap that limits the ability to measure progress on optimization itself. Existing benchmarks for trip planning provide reference itineraries rather than definitive ground truth, which means they cannot distinguish an optimal plan from a merely feasible one~\cite{chaudhuri2025tripcraft,shen2025triptailor}. As a result, evaluation only relies on language models acting as judges, an approach that introduces subjectivity and bias, or some feasibility metrics. In addition, current benchmarks do not introduce the detailed question categories for different complexities, which makes it hard to analyze where a system’s optimization process fails. Without clearly defined question categories, the resulting evaluation remains coarse and hard to interpret. There, it is impossible to quantify how well a system performs optimization currently.

To address the methodological gap, we adopt an \emph{agentic AI} framework that is designed for adaptive reasoning and self-correction during inference. Instead of following a fixed workflow in previous single and multiple agent systems~\cite{li2024survey}, our system is organized around a set of collaborating agents that can revise assumptions and recover from intermediate failures. At the center of this architecture is an \textit{Orchestration Agent} that manages subtask decomposition and oversees iterative refinement through explicit re-thinking cycles. It coordinates with an \textit{In-Vehicle Agent} that interfaces with user inputs and onboard conditions, as well as a \textit{Pool of Specialized Agents} responsible for domain-specific knowledge such as traffic, charging, and points of interest. This design enables the system to reason about both the problem and its own intermediate outputs, providing the flexibility and robustness required for multi-constraint optimization in intelligent vehicle scenarios.

To address the evaluation gap, we introduce the \textbf{T}rip-planning \textbf{O}ptimization \textbf{P}roblems Dataset (TOP), which is designed specifically for evaluating optimization in trip planning. Unlike existing benchmarks that rely on reference answers or subjective judging, TOP provides definitive ground truth for each query, enabling objective assessment of whether a system identifies the optimal solution under the stated constraints. The benchmark also organizes queries into well-defined reasoning categories, making it possible to analyze performance at a more granular level rather than relying solely on an end-to-end score. By combining objective optimality labels with category-level structure, the TOP Benchmark offers a clearer view of how and where optimization systems fail and enables more reliable comparison across different planning methods. Overall, we summarize our contribution as:


\begin{itemize}[noitemsep,leftmargin=*]
    \item We propose a hierarchical \textit{Agentic AI System} that explicitly models orchestration, collaboration, and self-correction among specialized agents for vehicle-centric trip planning.
    \item We construct the \textit{TOP Benchmark}, a dataset containing 500 queries across 15 reasoning categories and three difficulty levels, designed to test multi-constraint reasoning, temporal adaptation, and preference handling.
    \item We demonstrate that our system achieves a substantial improvement of \textit{77.4\% overall accuracy} compared to 30.4\% for single-agent and 23.6\% for workflow-based multi-agent baselines, highlighting that orchestrated agentic design is key to a robust trip planning optimization.
\end{itemize}

\section{Definition}

\begin{figure}[t]
    \centering
    \includegraphics[width=0.5\textwidth]{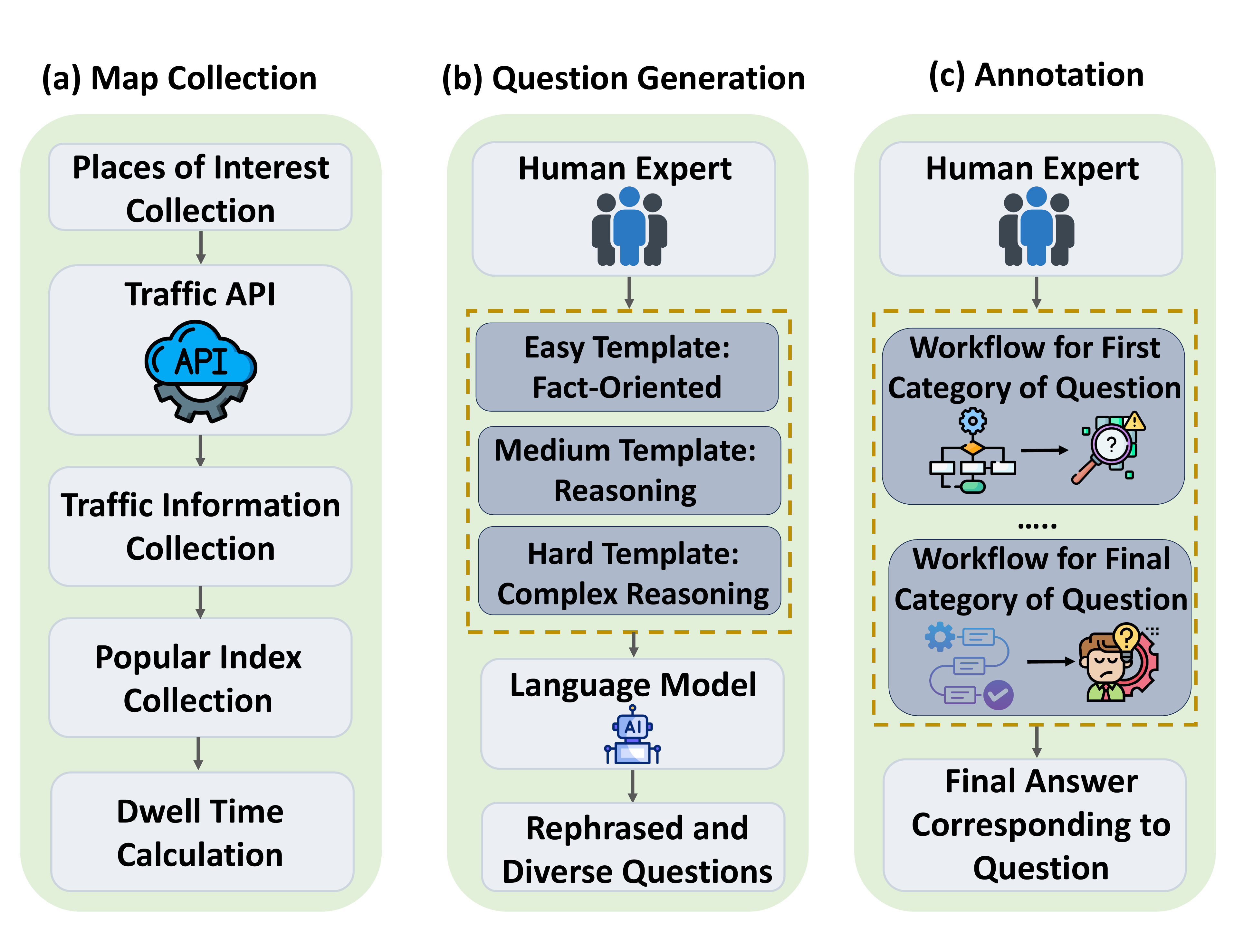}
    \caption{Dataset Generation Pipeline for our TOP dataset. In detail, our dataset construction contains three parts, and our annotation part ensures that each question will have a ground truth for evaluation.}
    \label{fig:dataset}
\end{figure}

We formalize the trip planning task under the Agentic AI paradigm, introducing notation, communication structure, and reasoning dynamics tailored for in-vehicle decision intelligence.

Let $Q$ denote a natural language query from a user or in-vehicle interface, expressing explicit requirements (e.g., ``I need to charge the vehicle and buy coffee before going to work'') and implicit preferences (e.g., preferred brand, minimal detour).  
Let $\mathcal{P} = \{p_1, p_2, \dots, p_n\}$ be the set of Points of Interest (POIs), where each POI $p_i$ is represented as a tuple:
$p_i = (x_i, y_i, c_i, \tau_i),$
with $(x_i, y_i)$ denoting spatial coordinates, $c_i$ denoting the category (e.g., café, charging station), and $\tau_i$ denoting the expected dwell time.

\begin{problem}[Trip Planning]
The goal of the trip planning is to construct an ordered itinerary:
$
I = \langle p_{i_1}, p_{i_2}, \dots, p_{i_T} \rangle,
$
that minimizes the total trip cost
$\mathcal{L}(I; Q, \mathcal{P}).$ 
\end{problem}

Typically, the cost function is defined as: $\mathcal{L}(I; Q, \mathcal{P}) = 
\sum_{t=1}^{T-1} 
\Big(
\text{Travel}(p_{i_t}, p_{i_{t+1}}) 
+ 
\text{Dwell}(p_{i_t})
\Big),
$
which jointly captures travel time between successive locations and the expected dwell time at each stop in the planning.  
Here, $\text{Travel}(\cdot)$ represents the traffic-aware travel time between two POIs, while $\text{Dwell}(\cdot)$ follows the category- and popularity-dependent model defined in Section~\ref{sec:dataset}.  

The itinerary must also satisfy several feasibility constraints: 
(1) all required categories mentioned in $Q$ are included in the plan;  
(2) temporal feasibility and POI opening hours are respected; and  
(3) user preferences and vehicle state constraints (e.g., battery state-of-charge, range) are met.  The optimal plan $I^*$ is defined as the itinerary that minimizes $\mathcal{L}(I; Q, \mathcal{P})$ while satisfying these constraints. While real-world trip planning involves multi-objective trade-offs (e.g., time vs. energy consumption), we formulate the problem as a constrained single-objective optimization (minimizing time) to establish a deterministic ground truth for benchmarking.

To solve this problem, we model the Agentic AI System (AAS) as a collection of interacting agents $\mathcal{A} = \{A_1, A_2, \dots, A_m\}$ connected through a communication topology $\mathcal{C}$.  
Each agent $A_j$ is defined by its role description $R_j$, prompt specification $S_j$, accessible toolset $T_j$, and local memory $M_j$.  
Agents exchange structured messages that carry intermediate reasoning results, factual lookups, or validation signals.  
At the system level, the input to the AAS is the pair $(Q, \mathcal{P})$, consisting of the user query and the available set of POIs.  
The agents collaboratively process this input by iteratively (1) interpreting the intent of $Q$, (2) retrieving relevant contextual information from $\mathcal{P}$ and external tools, and (3) assembling partial subplans into a complete itinerary $I$.  
The final output of the system is the optimized itinerary $I^*$ that minimizes $\mathcal{L}(I; Q, \mathcal{P})$ under the given constraints. Detailed specification for the system can be found later in Section~\ref{sec:method}.

\section{Dataset Generation}

\label{sec:dataset}

To rigorously evaluate our proposed Agentic AI system for trip planning optimization, we construct a new dataset that reflects the complexity of real-world urban mobility scenarios. Existing benchmarks largely focus on feasible plans, which are insufficient to stress-test multi-agent reasoning under multiple constraints~\cite{wang2025survey}. Our dataset fills this gap by introducing a diverse collection of queries paired with well-defined answers, enabling systematic evaluation of itinerary construction, preference handling, and multi-step reasoning. Overall, the dataset contains 50 unique Points of Interest (POIs), covering residential, commercial, and recreational categories. From these POIs, we generate 500 question–answer pairs, spanning more than 15 unique categories of reasoning tasks. Each entry in the dataset is annotated with a deterministic ground-truth solution, ensuring reproducibility and reliability in the evaluation of trip planning optimization.

\subsection{Map Collection}
The selection of POIs in our dataset is motivated by the need to cover the full spectrum of daily urban mobility for trip planning. Specifically, we include residential places (apartments), workplaces (companies), and a variety of functional or recreational locations such as cafes, restaurants, gyms, charging stations, and markets. This diversity is essential to ensure that the dataset can represent realistic trip planning scenarios that combine commuting, errands, and leisure activities within a single itinerary. The detailed distribution of the POIs can be found in \cref{tab:poi_distribution}.

\begin{table}[t]
\centering
\caption{Distribution of POIs in our dataset.}
\begin{tabular}{l cc}
\toprule
\textbf{Category} & \textbf{Number of POIs} &\textbf{Base Time} \\
\midrule
Apartments        & 6 & N/A \\
Companies         & 5 & N/A \\
Charging Stations & 6  & 30 Minutes \\
Cafés             & 9  & 5 Minutes\\
Gyms              & 6  & 25 Minutes\\
Markets           & 7  & 20 Minutes\\
Restaurants       & 11 & 60 Minutes\\
\midrule
\textbf{Total}    & 50 & N/A  \\
\bottomrule
\end{tabular}
\vspace{-2mm}
\label{tab:poi_distribution}
\end{table}

Beyond the POIs themselves, we also collect detailed traffic data such as driving time, preparing for trip planning optimization. In practice, real-world systems may query traffic information in real time from navigation services. However, to guarantee reproducibility across evaluations, we instead store every data offline. In detail, we rely on the map service API to retrieve all pairwise travel times and distances between POIs. For each origin–destination pair, we query the API under four representative time buckets (\textit{09:00}, \textit{12:00}, \textit{18:00}, and \textit{00:00}), thereby capturing rush-hour congestion as well as off-peak conditions. This design enables models to reason not only about spatial constraints but also about the temporal variability of travel costs for optimal trip planning.

\begin{table*}[t]
\centering
\caption{Overview of question categories across different difficulty levels. Each category is realized through a fixed workflow that deterministically produces the ground-truth answer.}
\begin{tabular}{p{2cm} p{4cm} p{9.5cm}}
\toprule
\textbf{Level} & \textbf{Category} & \textbf{Description} \\
\midrule
\multirow{6}{*}{Easy} 
& Name Lookup & Retrieve the name of all POIs within a given category. \\
& Travel Time (Driving) & Compute driving time between two specified locations. \\
& Travel Time (Walking) & Compute walking time between two specified locations. \\
& Distance Query & Return the pairwise distance between two POIs. \\
& Dwell Time Lookup & Return the expected dwell time of a POI based on its category and popularity. \\
& Nearest Neighbor Search & Identify the closest POI of a given type to a specified origin. \\
\midrule
\multirow{5}{*}{Medium} 
& Plan Evaluation & Given a fixed itinerary, calculate its total travel and dwell time. \\
& Route Comparison & Compare two alternative itineraries and identify the shorter one. \\
& Contextual Recommendation & Recommend an intermediate stop that best fits into an ongoing plan. \\
& Temporal Optimization & Select the optimal departure time that minimizes trip duration. \\
& Single-Factor Optimization & Construct or assess a plan while optimizing for a single dimension (e.g., cost, dwell time, or category preference). \\
\midrule
\multirow{5}{*}{Hard} 
& Full Itinerary Construction & Build a complete multi-stop trip plan from scratch. \\
& Multi-Constraint Planning & Generate itineraries that satisfy multiple explicit requirements simultaneously. \\
& Preference-Aware Planning & Construct plans that respect user preferences over categories or brands. \\
& Custom Dwell-Time Planning & Incorporate user-specified dwell times into itinerary generation. \\
& All-Intention Planning & Solve complex queries requiring the system to fulfill all possible intentions within a single coherent plan. \\
\bottomrule
\end{tabular}
\label{tab:question_categories}
\end{table*}

In addition to traffic, we also include the popular index in the data, which reflects the expected crowd level at different times of day. The popular index values are again derived from map service data and represent relative busyness across the weekday cycle. We integrate these indices with category-specific dwell-time models to simulate realistic activity durations. For example, restaurants exhibit longer stays and higher variance during meal times, while gyms and markets follow different temporal patterns. All the detailed time is adjusted based on the popular time as well as the base time in the category. Concretely, if no waiting occurs, the dwell time equals the category-specific base time $B_c$; otherwise, it becomes
\[
D(t,c) = B_c + \frac{p(t)}{100}\cdot B_c,
\]
where $p(t)$ denotes the popularity index at time $t$. By combining traffic-aware travel times with popularity-driven dwell times, our dataset faithfully models both mobility and activity dynamics, creating a challenging and realistic benchmark for agentic trip planning systems.

\subsection{Question Generation with Difficulty Levels}

To capture a spectrum of reasoning complexity, we organize the dataset into three difficulty levels: \textit{easy}, \textit{medium}, and \textit{hard}. The design ensures that each level increases the reasoning burden in a controlled and interpretable manner. However, for optimal trip planning, we always ask for the shortest time trip plan under constraints for all questions related to trip planning. 

\paragraph{Easy level}
The simplest level consists of fact-oriented queries that can be resolved by a single reasoning step. 
Typical tasks include asking for the travel time or distance between two specific locations, identifying the dwell time at a single point of interest given its category, or listing all available candidates within a certain type. 
These queries primarily test factual correctness and basic information retrieval.

\paragraph{Medium level}
The medium level introduces compositional reasoning across multiple steps. The questions involve optimal itineraries with two to three stops, where the system must calculate the total duration by combining travel and dwell times. 
Other representative tasks include comparing two alternative routes to determine which is more efficient, recommending the most suitable stop within a given category for the best trip planning, or finding the optimal departure time that minimizes total trip cost. 
Although each query still focuses on a single objective, solving them requires a chaining of multiple reasoning components.

\paragraph{Hard level}
The hardest level contains queries that demand multi-intention reasoning under simultaneous constraints. 
For example, the system may be asked to construct a trip that satisfies several goals—such as visiting a café, a gym, and a market—while also respecting time budgets and user preferences (e.g., preferring a specific brand of gas station or limiting dwell time at a restaurant). 
These scenarios often require enumerating feasible plans and reasoning over trade-offs between competing objectives to get the optimal plan. 
They represent the most realistic and challenging cases in trip planning, pushing systems beyond rigid workflows towards adaptive, agentic reasoning.

\paragraph{Implementation}
For each difficulty level, we predefine a closed set of question categories, each realized by a human-written prompt template. For the easy level, we predefine 6 unique question categories, while we predefine 5 categories for medium and hard, respectively. The prompt template includes a fixed string with slots for places and times. At generation time, we instantiate a question by randomly sampling eligible POIs from the category-specific pools and filling the template slots. Sampling enforces the uniqueness of intermediate and final stops by re-drawing until a new ID is obtained, thereby avoiding duplicates. To preserve validity, we apply structural constraints during category sampling (e.g., forbidding the co-occurrence of gas and charging). Times are also randomly sampled from the available time slots. Finally, after we get the initial fixed template, we apply Gemini-2.5-pro~\cite{comanici2025gemini} to paraphrase to improve linguistic variety while preserving semantics. In total, we generate 100 questions for the easy level, 200 questions for the medium level, and 200 for the hard level.

\begin{figure*}[t]
    \centering
    \includegraphics[width=1.0\textwidth]{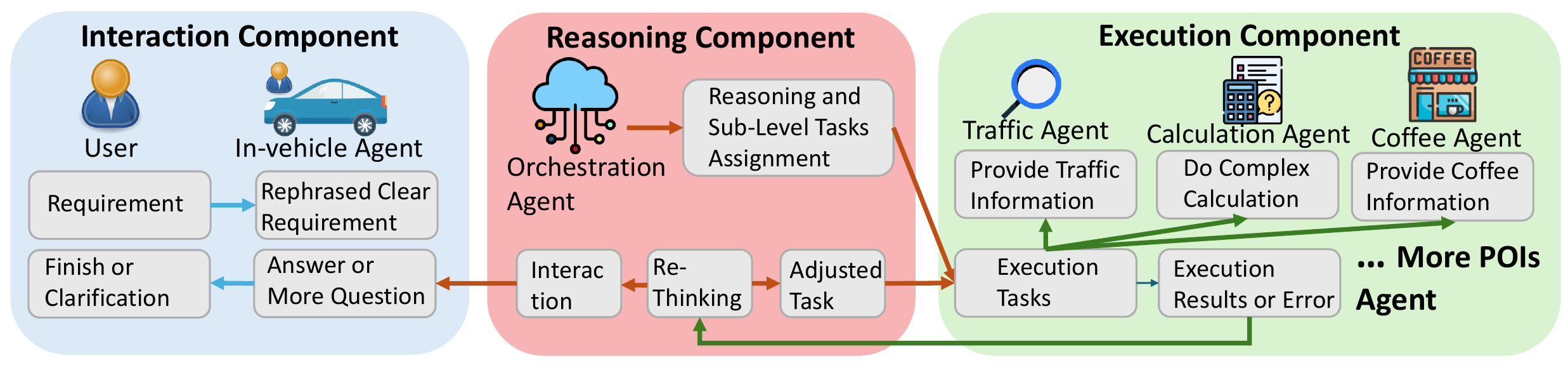}
    \vspace{-2mm}
    \caption{Overview of our Agentic AI system. Our system contains three different components. The interaction component ensures a smooth conversation with the user, the reasoning component will divide the task into sub-level tasks and reflect the plan, and the execution component will execute the tasks from the reasoning component. All three components work collaboratively to ensure the optimization of trip planning.}
    \label{fig:Agent_Archi}
    \vspace{-4mm}
\end{figure*}

\subsection{Annotation}

To ensure the reliability and objectivity of our TPO dataset, every query in the dataset is paired with a deterministically generated ground-truth solution. Instead of relying on manual annotation, which is nearly impossible due to the complexity of the problem, we design explicit pipelines to obtain the ground-truth answer, which ensures reproducibility. 

At the core of this process, each question template is associated with a predefined \textbf{workflow} that encodes how to compute its answer. For questions at the easy level, the workflow directly queries stored travel distances, traffic-adjusted durations, or category-specific dwell-time functions. For harder queries, the workflow composes multiple operations, such as simulating an itinerary, accumulating dwell and travel times, or enumerating itineraries under multiple constraints. When user preferences or custom dwell times are specified, the workflow explicitly incorporates these constraints before computing the final solution to ensure all labels meet requirements. Note that the enumeration of itineraries is employed solely for the offline construction of the dataset, and our proposed Agentic AI framework (Section III) does not require enumeration.

While these workflows ensure deterministic supervision, it is important to note that they cannot by themselves serve as a general trip planning solution. Each workflow is rigidly tied to a predefined question template, requiring exact matches between the input format and the underlying database schema. As a result, workflows cannot interpret natural language variations, resolve ambiguities, or generalize to unseen formulations of user goals. Besides, workflows are also highly reliant on human effort, which costs a lot of time.

To preserve clarity, we separate answers into primary and auxiliary outputs. For instance, in multi-objective questions, the workflow returns both the best itinerary and the reasoning behind its selection for auxiliary outputs, such as the total time. Finally, all answers are formatted in a structured JSON schema, making them easy to evaluate programmatically. This workflow-based annotation strategy provides deterministic, verifiable ground truth for every query, ensuring that evaluation results are both reliable and reproducible.

\section{Agentic System for Trip Planning}
\label{sec:method}

In this section, we introduce our Agentic AI system as well as the method that improves the performance for the trip planning optimization task.

\subsection{Architectural Overview}
Our framework is organized into three principal components: a user-facing In-Vehicle Agent for \textbf{interaction component}, a central Orchestration Agent for \textbf{reasoning component}, and a decentralized Pool of Specialized Agents for \textbf{execution component}. The Orchestration Agent serves as the strategic nucleus of the system, receiving contextualized queries from the In-Vehicle Agent and coordinating the actions of the specialized agents to fulfill the user’s goals. This hierarchical design centralizes high-level reasoning while distributing domain-specific computations, ensuring both coherence and efficiency. It is worth noting that while our current cost function focuses on travel and dwell time, the proposed agentic framework is inherently extensible.

\subsection{Agent Specializations}
The framework’s power derives from the distinct but complementary roles of its agents. Together, they collaborate to solve complex, multi-faceted planning tasks.

\paragraph{\textbf{Interaction Component}}
Interaction Component contains the \textbf{In-Vehicle Agent}, which is the user-facing agent and the first entry point of the system. It parses natural language instructions and transforms them into structured queries. By grounding user intent in contextual signals, this agent reduces ambiguity and ensures the orchestration layer receives well-formed, actionable goals. Besides, it will also transfer the final answer or a potential further clarification question to the user for interaction.

\paragraph{\textbf{Reasoning Component}}
Reasoning Component contains \textbf{Orchestration Agent}, functioning as the central controller, the Orchestration Agent governs the entire planning lifecycle. It decomposes high-level goals into sub-tasks, selects the appropriate specialized agents and dispatches tasks to them, and integrates the results into a coherent itinerary. Importantly, it is explicitly instructed to perform \emph{re-thinking} when inconsistencies or infeasibility are detected, ensuring robustness beyond static workflows.

\paragraph{\textbf{Execution Component}}
Execution Component contains a specialized agent pool, which embodies domain expertise. Its members include:

\textbf{Traffic Agent:} An information-retrieval agent with access to the traffic information. It provides travel times and distances between POIs conditioned on the time.  

\textbf{Calculation Agent:} The optimization engine responsible for aggregating travel and dwell times to score itinerary costs. Beyond sequential aggregation, it reasons about concurrent efficiencies—for instance, determining whether a café visit suffices to cover an EV charging requirement.  

\textbf{Point-of-Interest (POI) Agents:} Inspired by the smart-city paradigm~\cite{roscia2013smart,kalyuzhnaya2025llm}, each POI category is represented by a dedicated agent (e.g., Coffee Agent, Gym Agent). These agents act as digital twins of their physical counterparts, returning expected dwell times according to the popularity-driven model $D(t,c)$, along with attributes such as wait times, operational hours, and monetary cost.

\begin{figure}[t]
    \centering
    \includegraphics[width=0.5\textwidth]{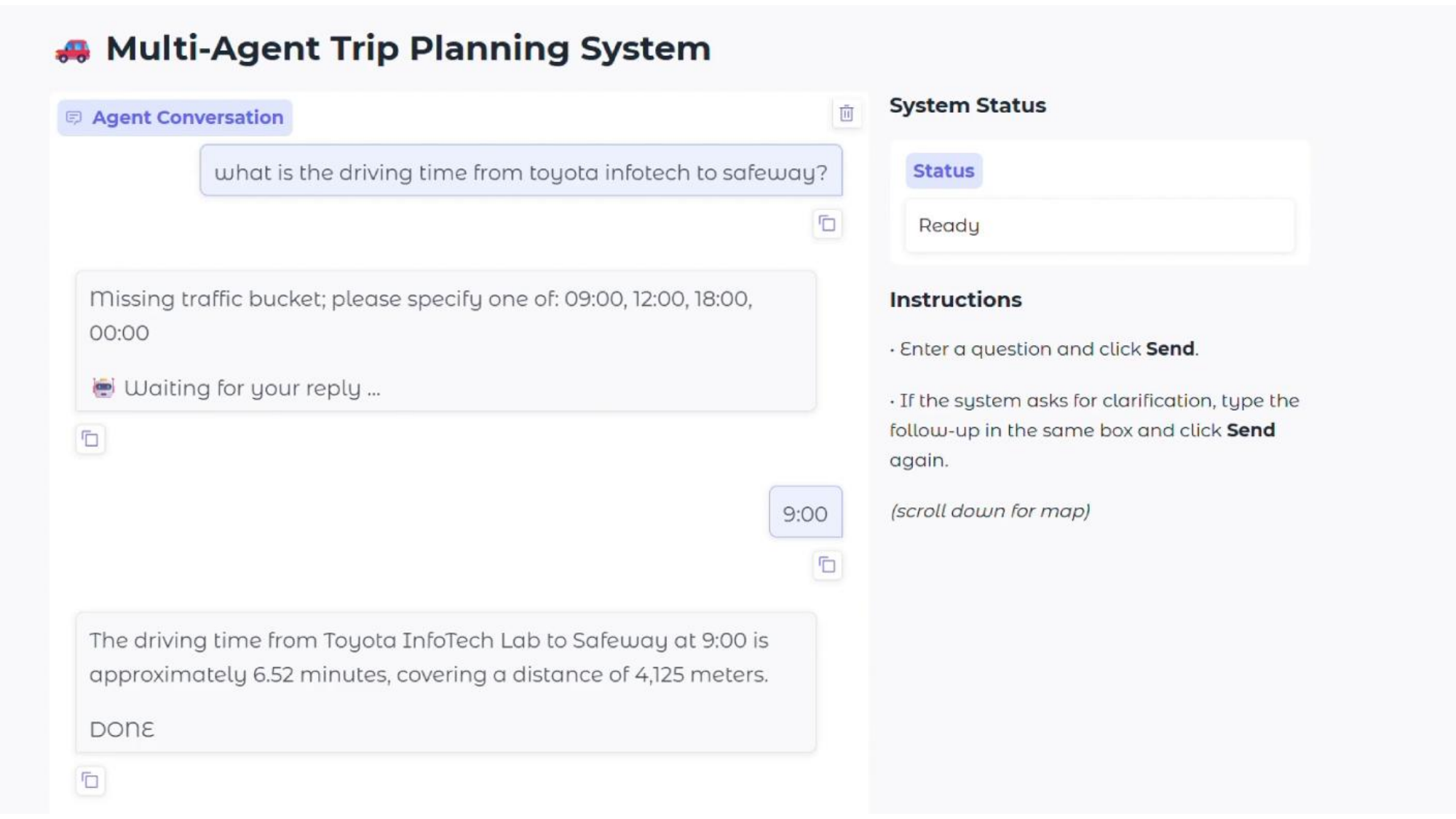}
    \caption{An interaction example for our system with a graphical user interface. Though our system could interact with users, we discard the interactive ability when compared with baselines to ensure a fair comparison.}
    \vspace{-5mm}
    \label{fig:GUI}
\end{figure}

\subsection{Collaborative Planning Protocol}

The agents collaborate through a structured protocol managed and directed by the Orchestration Agent. The process for solving a complex query is not a rigid chain but a dynamic workflow. The protocol is initiated when the In-Vehicle Agent captures and relays a user query to the Orchestration Agent. Upon receipt, the Orchestrator analyzes the user's intent and decomposes the goal into a strategic plan composed of interdependent sub-tasks. Subsequently, the Orchestrator delegates these sub-tasks to the appropriate specialized agents. The information gathered from these experts—containing candidate POIs such as their expected dwell times is then synthesized and returned to the Orchestrator. And Orchestrator will decide the next step accordingly based on the information. For example, the Orchestrator will rethink the action if it receives some error information or reevaluate the plan to ensure the plan meets the requirements of users.

\section{Experimental Evaluation}

In this section, we provide our experimental results to show the effectiveness of our proposed agentic method.

\begin{table}[h!]
\centering
\caption{Performance comparison across difficulty levels in TOP on our proposed agent system and two baselines. The results show that our proposed agentic system consistently performs the best among all methods, showing the effectiveness of our method.}
\label{tab:performance_comparison}
\begin{tabular}{lccc}
\toprule
\textbf{Query Difficulty} & \textbf{Ours} & \textbf{Swarm} & \textbf{Single-Agent} \\
\midrule
Easy & 97\% & 71\% & 82\% \\
Medium & 87\% & 15\% & 26\% \\
Hard & 58\% & 8.5\% & 9\% \\
\midrule
\textbf{Overall} & \textbf{77.4\%} & \textbf{23.6\%} & \textbf{30.4\%} \\
\bottomrule
\end{tabular}
\end{table}

\begin{figure*}[t]
    \centering
    \includegraphics[width=0.8\textwidth]{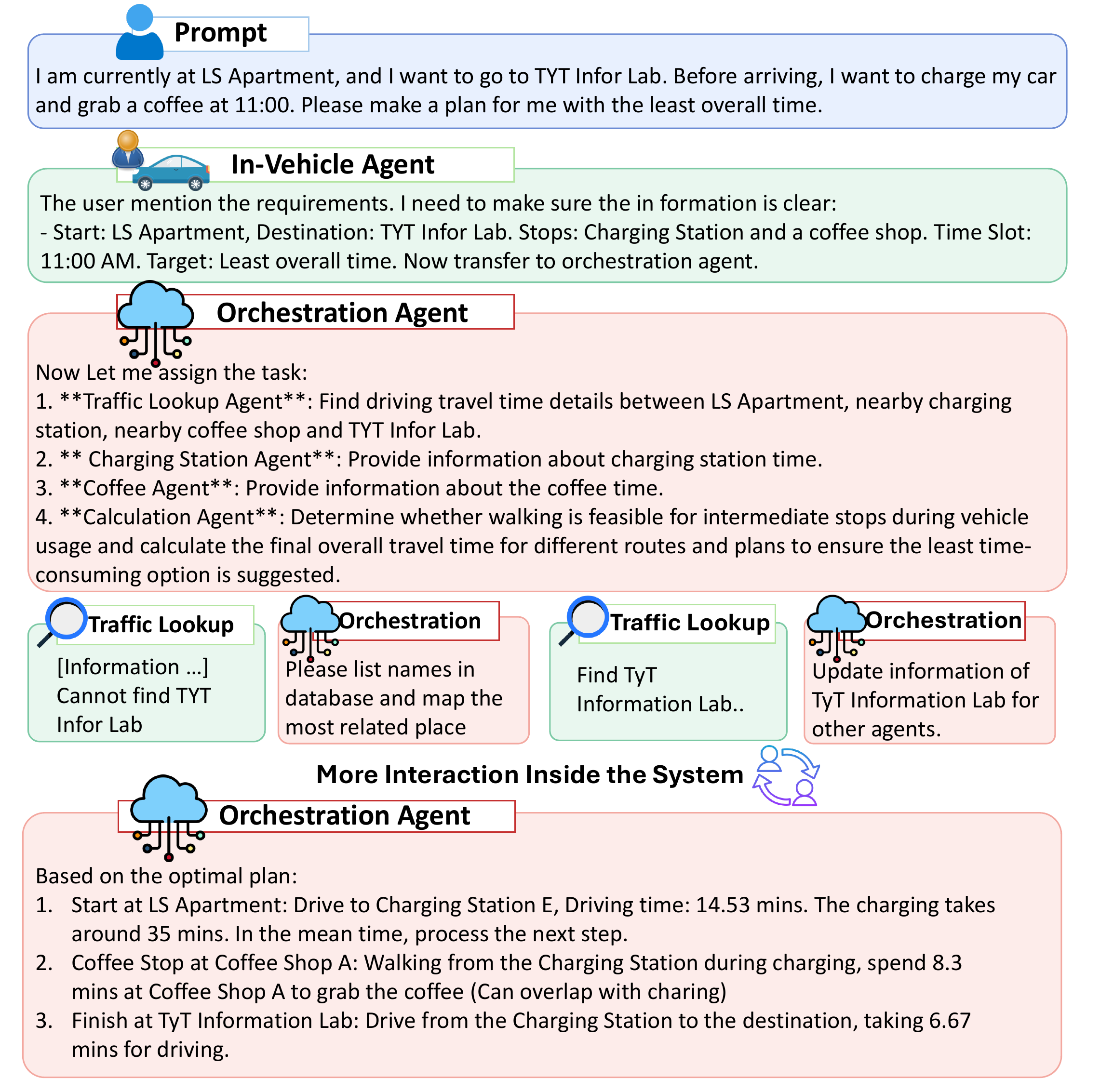}
    \caption{Case used in our analysis. In this case, our system shows that it is possible to correct the error by the system itself.}
    \label{fig:case}
    \vspace{-5mm}
\end{figure*}

\subsection{Evaluation Setting}
\noindent \textbf{Setup Overview.}
We evaluate all models on the TOP benchmark introduced in ~\cref{sec:dataset}, which covers diverse urban planning queries with deterministic ground-truth solutions. All experiments are conducted under identical map, traffic, and popularity conditions to ensure the correctness of the evaluation.

\noindent \textbf{Baselines}
To validate the effectiveness of our agentic framework, we compare the performance of our proposed Agentic AI system against two key baselines that represent alternative approaches discussed in our introduction:
\begin{itemize}
\item \textbf{Single-Agent LLM:} This baseline utilizes a single Large Language Model (LLM), which we argue often struggles with the multi-step reasoning required for complex itinerary planning optimization. The agent is given the same user query and access to the same set of tools as our system, but must solve the problem without the collaborative, multi-agent structure.
\item \textbf{SWARM Framework:} This baseline follows the SWARM design pattern, where agents collaborate via handoffs: each agent can locally decide to delegate the conversation to a more suitable peer while sharing the same message context. Instead of massive parallel sampling, SWARM emphasizes lightweight, controllable multi-agent orchestration via routines and handoffs, enabling decentralized task routing without a central orchestrator~\cite{openai_swarm_cookbook_2024}.
\end{itemize}

\noindent \textbf{Evaluation Metrics}
The primary metric for our evaluation is \textbf{Accuracy}. An answer is considered correct only if the final itinerary generated by the system exactly matches the deterministically generated ground-truth solution provided in the dataset. This strict metric ensures that we are evaluating the correctness of the entire reasoning and planning process, from understanding user intent to optimizing the final plan.

\noindent \textbf{Implementation Details}
For our proposed Agentic AI system and other baseline systems, all agents are powered by the GPT-4o model~\cite{openai_hello_gpt4o_2024}. All systems are provided with the same set of tools and have access to the same offline database containing POI information, pairwise travel times, and popularity data to ensure a fair and reproducible comparison. As shown in \cref{fig:GUI}, our system is able to interact with users for clarification information. However, to have a fair comparison with baselines, we disable the ability of our system to interact with users in the following experiments. We implemented our multi-agent environment using the Microsoft Autogen \cite{wu2024autogen} framework, which provides structured tools for orchestrating agent interactions and managing communication workflows.

\subsection{Evaluation Results}
To show the effectiveness of our proposed agentic system in trip planning, we compare our proposed system with other baselines using our generated dataset TOP. In detail, for every question in the dataset, we allow each system only one trial and calculate the accuracy based on the difficulty levels. In detail, the results are presented in \cref{tab:performance_comparison}, and we can have the following observation:

\noindent$~\bullet$ \textbf{Superior Overall Performance and Architectural Advantage:}  Our agentic framework achieves the highest overall accuracy of \textbf{77.4\%}, substantially outperforming both the Single-Agent (\textbf{30.4\%}) and Swarm (\textbf{23.6\%}) baselines. This demonstrates that centralized orchestration with specialized sub-agents leads to more reliable and consistent reasoning.

\noindent$~\bullet$ \textbf{Robustness to Increasing Task Complexity:} The most significant performance gap emerges as task difficulty increases. While all systems perform adequately on "Easy" queries, the baselines experience a dramatic collapse on "Medium" and "Hard" tasks. Our system maintains a high 87\% accuracy on medium-level compositional reasoning tasks, while the baselines drop to 15-26\%. This robustness extends to the "Hard" queries, which require multi-intention reasoning under simultaneous constraints. Here, our system's 58\% accuracy is nearly seven times higher than the baselines. This demonstrates that our architecture is uniquely capable of managing the complex, multi-step, and interdependent reasoning required for realistic trip planning, whereas the baselines prove almost entirely incapable of solving these complex problems.

\noindent$~\bullet$ \textbf{Single-Agent vs. Multi-Agent:} An important observation from our experiments is that the superiority of a multi-agent system is \emph{not guaranteed by default}.  Although multi-agent frameworks are conceptually more powerful due to distributed reasoning and specialization, poorly coordinated systems can even underperform a single-agent baseline. For example, in our results, the Single-Agent model outperforms the generic multi-agent Swarm framework. This finding highlights that simply deploying multiple agents without explicit orchestration or task structure may lead to communication overhead, redundant reasoning, or inconsistent outputs.

In contrast, our proposed agentic system demonstrates that when multi-agent collaboration is well designed and strategically orchestrated, it can achieve significant gains across all difficulty levels. The Orchestration Agent in our architecture acts as a central meta-controller that manages sub-task decomposition, enforces consistency of the reasoning, and triggers re-thinking when necessary. This structured collaboration enables agents to reason coherently and adaptively rather than acting as isolated entities.



\subsection{Case Study}

To concretely demonstrate our system's robustness, we analyze a representative case from the Hard difficulty tier, with its detailed flow illustrated in \cref{fig:case}. The case originates from a complex, multi-intention query: a user requests a plan from LS Apartment to TYT Infor Lab, with intermediate stops for car charging and coffee. The plan must optimize for the least overall time at 11:00 AM.

During the initial planning phase, the system encounters a critical failure. The Traffic Agent reports an error, as it cannot find the user-specified destination TYT Infor Lab due to a data mismatch with the POI database. This event immediately triggers our architecture's core advantage. The Orchestration Agent, instead of terminating the process, intercepts this sub-task failure and activates its re-thinking protocol. It diagnoses the issue as a likely entity resolution problem and dynamically generates a new instruction, directing the Traffic Agent to search for and map to the most related place name.

This capacity for centralized diagnosis and self-correction is crucial. In contrast, a decentralized framework like Swarm, which lacks a top-level coordinator to diagnose the root cause of a sub-task failure, would likely become trapped in a repeat generation loop, repeatedly re-issuing the failed query until the entire task times out.

After successfully resolving the location error, the system proceeds to the optimization phase. The Orchestration Agent coordinates the various specialized agents and, through reasoning performed by the Calculation Agent, identifies a significant opportunity for concurrent optimization. The final, optimal plan instructs the user to walk to the coffee shop and complete that task during the vehicle's 35-minute charging window. This case clearly demonstrates our architecture's ability to not only use agentic re-thinking to overcome real-world ambiguities and errors but also to execute complex, multi-constraint reasoning to discover optimized solutions.

\section{Related Work}

There are other works that focus on the trip planning, while they differ significantly from our work. For example, Xie et al. ~\cite{xie2024travelplanner} proposes TravlerPlanner, a benchmark for travel planning. Shen et al.~\cite{shen2025triptailor} propose a larger dataset with more POIs compared with Travelplanner, and Chaudhuri~\cite{chaudhuri2025tripcraft} also propose another benchmark for travel planning, which provides more fine-grained constraints. However, all of these datasets do not provide definitive ground-truth labels, but only reference answers for evaluation, which makes the evaluation process highly relies on LLM-as-a-Judge that could potentially introduce bias. Moreover, all previous datasets lack question categories, making it hard to perform a fine-grained diagnosis of where an agentic system fails. For solutions to the travel-planning problem, while operations research methods~\cite{kucukoglu2021electric,schneider2014electric} offer exact solutions for mathematically defined problems, they struggle with unstructured natural language inputs. However, future works might include using LLMs to use these solvers. Li et al.~\cite{li2024personal} proposes an agent that tries to solve TravlerPlanner. Besides, Fang et al.~\cite{fang2024travellm} and Tane et al.~\cite{tang2024itinera} also propose LLM-based agentic systems for trip planning based on public transportation or walking. However, most of the previous works are based on the single-agent system, while our paper makes a pioneering effort in a multi-agent system for trip planning. Besides, most previous works aim at generating a feasible plan while ours always aim at the optimal trip planning.

\section{Conclusion}
In this work, we address both the methodological and evaluation gaps inherent in the trip planning optimization task by jointly proposing a hierarchical agentic AI system and the TOP benchmark. To close the methodological gap, we develop a centrally orchestrated multi-agent framework that supports adaptive reasoning, sub-task decomposition, and self-correction, enabling the system to optimize itineraries under multiple constraints rather than simply generating feasible plans. To close the evaluation gap, we introduce the TOP benchmark, which provides deterministic optimal solutions by fixed workflows designed by humans and a structured set of reasoning categories across multiple difficulty levels, allowing objective and fine-grained assessment of optimization performance. Experimental results demonstrate that our system achieves 77.4\% accuracy and significantly outperforms single-agent and workflow-based multi-agent baselines, particularly on medium and hard tasks, highlighting the importance of structured orchestration and agentic re-thinking in solving complex optimization problems. Overall, our work establishes a solid foundation for future research on agentic AI systems capable of robust, interpretable, and high-performance trip planning optimization.


\bibliographystyle{IEEEtran}
\bibliography{biblog}

\end{document}